\documentclass{article}
\makeatletter
\let\icasspArticleSect\@sect
\makeatother
\usepackage{spconf}
\makeatletter
\let\icasspConferenceSect\@sect
\def\@sect#1#2#3#4#5#6[#7]#8{%
  \ifnum#2>3\relax
    \icasspArticleSect{#1}{#2}{#3}{#4}{#5}{#6}[#7]{#8}%
  \else
    \icasspConferenceSect{#1}{#2}{#3}{#4}{#5}{#6}[#7]{#8}%
  \fi}
\makeatother
\usepackage[utf8]{inputenc}
\usepackage{times,soul,url}
\usepackage[protrusion=false]{microtype}
\usepackage[hidelinks]{hyperref}
\usepackage[small]{caption}
\usepackage{graphicx,amsmath,amssymb,amsthm,booktabs,algorithm}
\graphicspath{{figures/}}
\DeclareUnicodeCharacter{00D7}{\ensuremath{\times}}
\usepackage{algpseudocode,algorithmicx,tabularx,placeins,makecell}
\name{Ruiheng Wang$^{1}$, Yubo Hou$^{1}$, Yakun Zhu$^{2}$, Tianle Shen$^{2}$, Tao Wan$^{1}$, and Zengchang Qin$^{1,3,\dagger}$\footnotemark[1]}
\address{\small $^{1}$Beihang University, Beijing, China;  $^{2}$Shanghai Jiao Tong University, Shanghai, China; $^{3}$VinUniversity, Hanoi, Vietnam\\
\small pierrewang2024@gmail.com, \{taowan,zcqin\}@buaa.edu.cn}
\title{HeuFouFT: Task-Guided Metaheuristic Coordinate Search for Fourier Fine-Tuning}

\begin{document}
\microtypesetup{expansion=false}

\maketitle
% Footnotes on the first page: the dagger marks the corresponding author,
% and the code release link accompanies the availability note in the abstract.
\footnotetext[1]{$^{\dagger}$Corresponding author: zcqin@buaa.edu.cn.}
\footnotetext[2]{Code is available at \url{https://github.com/pierrewang2024-collab/HeuFouFT}.}

\begin{abstract}

We introduce \textbf{Heu}ristic-Guided \textbf{Fou}rier \textbf{F}ine-\textbf{T}uning (\textbf{HeuFouFT}), a task-guided framework for selecting trainable frequency coordinates in Fourier fine-tuning. Existing uniform and Gaussian band-pass schemes allocate a limited spectral budget through fixed, task-agnostic rules. HeuFouFT instead searches for coordinates using downstream performance. A coarse intensity map from lightweight block-level probes initializes three metaheuristic optimizers: Genetic Algorithm with Simulated Annealing (GA-SA), Particle Swarm Optimization (PSO), and Cuckoo Search (CS). During search, a Random Forest filters each population so that only the top 30\% of candidates proceed to proxy fine-tuning. On E2E with GPT-2-Medium, all three variants outperform random-uniform FourierFT, Gaussian band-pass FourierFT, and LoRA across five metrics. PSO further outperforms LoCA, the best-performing baseline, on four metrics while using 37.6\% fewer trainable spectral coefficients. Once coordinates are selected, HeuFouFT requires only 15--18\% FLOPs of Full FT. These results show that task-guided search allocates limited spectral capacity more effectively than fixed sampling. Our code is publicly available.\footnotemark[2]

\end{abstract}


\section{Introduction}

\begin{figure}[t]
    \centering
    % Spectrum of the LoRA-derived $\Delta W$ from GPT-2-Medium fine-tuned on E2E
    % (rank 8, layer-1 Q projection). Reproducible from
    % code/analysis/figure1_sampling_vs_spectrum.py --layout origin
    % --lora_adapter <run>/final_adapter.pt.
    \includegraphics[width=\columnwidth]{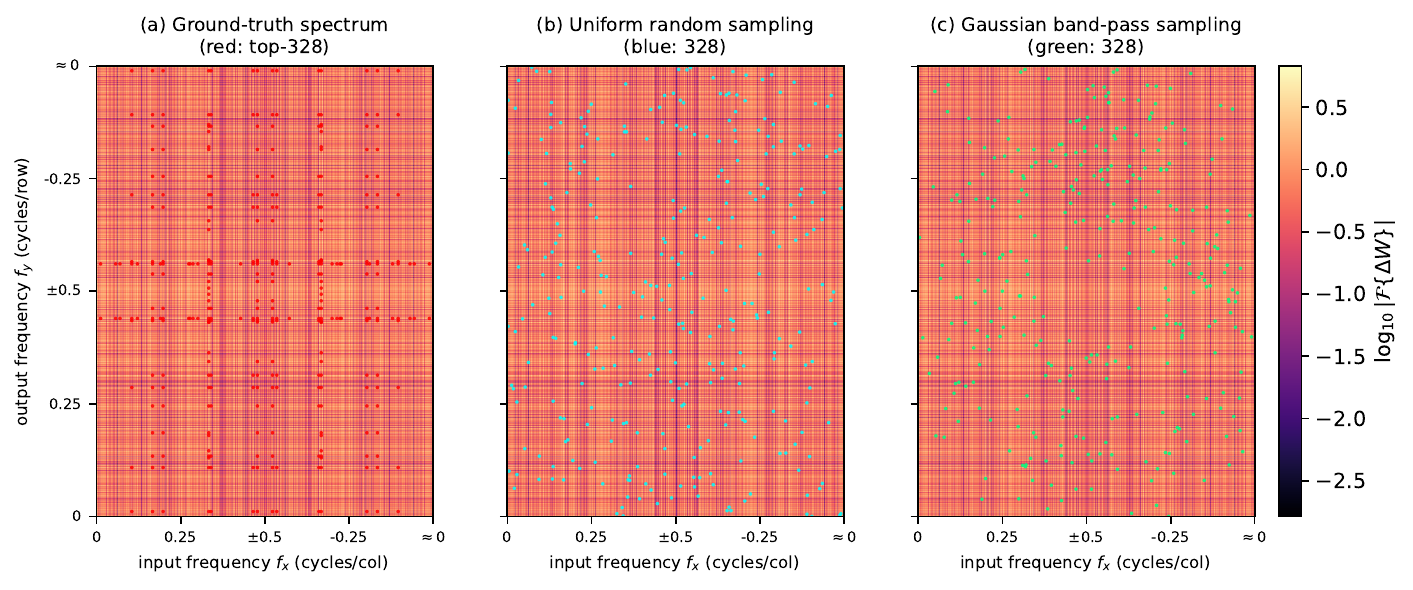}
    \caption{Log-magnitude spectrum of the layer-1 $Q$ projection $\Delta W$ from GPT-2-Medium LoRA fine-tuned on E2E (rank~8). Red markers indicate the 328 highest-energy frequency bins; blue markers show the 328 uniform random samples; green markers show 328 Gaussian band-pass samples.}
    \label{fig:gpt2_spectra}
\end{figure}

Parameter-efficient fine-tuning (PEFT) adapts large language models by learning a small set of parameters while keeping pretrained weights frozen. LoRA represents weight updates through low-rank factors~\cite{hu2021lora}. Spectral methods offer another parameterization by learning coefficients in a transform domain, where a compact representation can reduce the number of trainable parameters~\cite{xu2020learningfrequencydomain,gao2024parameter}.

Several approaches exploit spectral structure in adaptation. F-Adapter allocates adapter capacity according to spectral complexity~\cite{zhang2025fadapterfrequencyadaptiveparameterefficientfinetuning}, while FouRA applies Fourier and cosine transforms within low-rank adaptation for vision models~\cite{borse2024fourafourierlowrank}. LoCA jointly learns coefficients and their locations in a cosine representation~\cite{du2025loca}. HeuFouFT differs by selecting a discrete, layer-shared Fourier coordinate set before final fine-tuning through validation-guided derivative-free search.

FourierFT constructs weight updates through an inverse discrete Fourier transform of selected coefficients~\cite{gao2024parameter}. Its default uniform sampling and optional Gaussian band-pass sampling choose coordinates through fixed, task-agnostic rules, so the selected frequency coordinates cannot adapt to the downstream task.

The choice of coordinates matters when only a small number of Fourier components can be trained. As Figure~\ref{fig:gpt2_spectra} shows, a learned LoRA update can exhibit localized spectral energy that is poorly covered by uniform or radially constrained sampling. Although this spectrum is not an oracle for FourierFT, it motivates selecting coordinates using downstream task feedback.

To realize this selection, we draw on derivative-free optimization, which recent efforts \cite{jin2024derivativefreeoptimizationlowrankadaptation,qiu2025evolutionstrategiesscalellm} have applied to low-rank adaptation. Motivated by metaheuristic optimization~\cite{holland1975adaptation,kennedy1995particle,yang2009cuckoo}, we introduce \textbf{HeuFouFT}. Our contributions are threefold. First, we make Fourier coordinate selection \emph{responsive to downstream performance} rather than a fixed sampling rule. Second, we develop a \emph{surrogate-assisted framework} that combines a block-level task prior with three metaheuristic optimizers, GA-SA, PSO, and CS, to refine complete coordinate sets. Third, on GPT-2-Medium and E2E, all three variants outperform fixed sampling and LoRA across five metrics while using 37.6\% fewer spectral coefficients than the FourierFT and LoCA configurations; PSO further outperforms LoCA on four metrics.

\linespread{1.0}\selectfont
\section{Frequency-Domain Parameterization and Motivation}
\label{sec:fundamentals}

FourierFT parameterizes weight updates using a small set of Fourier components~\cite{gao2024parameter}. Given a frozen weight matrix $W_0 \in \mathbb{R}^{d_1\times d_2}$, let $E \in \mathbb{Z}^{2\times n}$ specify $n$ selected frequency coordinates and $c \in \mathbb{R}^{n}$ their trainable coefficients. The sparse spectrum $F(E,c)$ places $c_p$ at the coordinate in column $p$ of $E$, with zeros elsewhere. The weight update is
\begin{equation}
\Delta W = \alpha\,\Re\!\left(\mathcal{F}^{-1}\!\left(F(E,c)\right)\right),
\label{eq:forward}
\end{equation}
where $\mathcal{F}^{-1}$ denotes the inverse two-dimensional DFT, $\Re$ extracts the real part, and $\alpha$ controls the update scale. As in FourierFT, the coefficients are real scalars and the real projection in Eq.~\eqref{eq:forward} produces a real-valued update without requiring paired coefficients. Each layer learns its own $n$ coefficients while sharing $E$ and $\alpha$; the trainable parameter count is therefore exactly $n$ per adapted matrix.

For a fixed budget $n$, the coordinates in $E$ determine the Fourier components that can contribute to $\Delta W$. Learning $c$ adjusts their amplitudes within the resulting update subspace. \textbf{Frequency selection therefore controls the directions for adaptation.} Figure~\ref{fig:gpt2_spectra} reveals that LoRA-derived $\Delta W$ has its top energy bins concentrated in localized peaks, sharing little overlap with either uniform random sampling or Gaussian band-pass sampling. Fixed schemes cannot align their parameter budget with these irregular regions. Random-uniform FourierFT assigns every coordinate equal selection probability, while Gaussian band-pass sampling constrains selection to an annulus. Our search instead uses lightweight fine-tuning on coarse frequency blocks to initialize the optimizers with task-informed scores.

Spectral evidence alone cannot rank frequency regions by adaptation value. Figure~\ref{fig:gpt2_spectra} depicts one particular low-rank solution, and Table~\ref{tab4} shows that the preferred Gaussian band-pass setting varies across evaluation metrics. Task feedback is therefore needed to decide where the parameter budget should be placed, and the block-level probes introduced next provide this feedback.

\section{Frequency-Domain Sampling Approaches}\label{sec3}
We first consider two baseline strategies for selecting the frequency coordinates $E$. Uniform sampling selects $n$ points with equal probability, while Gaussian band-pass sampling favors an annulus controlled by frequency bias $f_c$ and bandwidth $W$. Setting $f_c=0$ gives a low-pass region.

\subsection{Heuristic Information Foundation}\label{sec3-2}
To guide sampling with task feedback, we partition the frequency domain into a $10\times10$ grid and probe each block through lightweight fine-tuning on 10\% of the training data for one epoch, scoring each probe on the validation split. Each probe returns the five NLG metrics, which we normalize and aggregate into a \emph{heuristic intensity}
\begin{equation}
\begin{aligned}
\mathrm{HI}={}&0.3\,\mathrm{BLEU}+0.2\,\mathrm{METEOR}\\
             &+0.2\,\mathrm{ROUGE\text{-}L}+0.15\,(\mathrm{NIST}+\mathrm{CIDEr}),
\end{aligned}
\label{eq25}
\end{equation}
where the normalization removes scale differences among metrics and the coefficients encode relative preferences over them~\cite{marler2010weighted}. For every spectral region, the \emph{intensity map} scores the downstream performance achievable by placing trainable coefficients there. Notably, the intensity distribution tracks the spectral energy concentration of the low-rank update in Figure~\ref{fig:gpt2_spectra}, with a Pearson coefficient of $r=0.85$ across the $10\times10$ frequency blocks, which supports the reliability of the heuristic map.

Instead of the spectral magnitude of the frozen model, this map measures adaptation value: a spectrally modest region may host coordinates that strongly benefit the task, and vice versa. Because each block score averages over an entire region, however, it cannot indicate which coordinates within a block matter. To resolve this, the metaheuristic searches start from the map, evaluate complete coordinate sets against task performance, and refine the selection below block granularity.

During search, a Random Forest acts as a filter for each newly generated population. It predicts candidate fitness, and only the top 30\% proceed to proxy fine-tuning and validation; the remaining 70\% are discarded without training. The surrogate achieves $R^2=0.86$ on held-out candidates. Its features, labels, and hyperparameters are documented in the released code.\footnotemark[2] All task feedback, including block probes, surrogate labels, and search fitness, is computed on the validation split; reported results use the held-out test set.

\subsection{Metaheuristic Sampling Algorithms}\label{sec3-3}

Starting from candidates initialized with the block-level prior, we compare three searches for the layer-shared coordinate set $E$ under a fixed budget on a $d\times d$ grid ($d=1024$). Each retained candidate is evaluated by fine-tuning its coefficients on 10\% of the training data for one epoch. Its validation HI score from Eq.~\eqref{eq25} serves as the search fitness.

\paragraph{Genetic-Simulated Annealing (GA-SA)} combines evolutionary search with probabilistic acceptance to reduce premature convergence~\cite{holland1975adaptation,bisht2004hybrid,liang2021hybrid}. Each chromosome encodes a complete set of frequency coordinates. Two-point crossover and bit-flip mutation ($p_m=0.05$) generate candidates. SA accepts candidates of equal or higher fitness, and accepts a decrease of magnitude $\delta$ with probability $e^{-\delta/T}$, allowing exploration beyond local optima. The temperature starts at $T=1000$ and decays by a factor of 0.95 per iteration.

\paragraph{Particle Swarm Optimization (PSO)} treats each sampling set as a particle whose coordinate velocities follow its personal best and the swarm's best solution~\cite{kennedy1995particle,oucif2025discrete,liu2025q}. The inertia weight decreases linearly from 0.9 to 0.4 over the first 200 iterations, remaining at 0.4 thereafter. For either integer coordinate $z$, the updated velocity $v^{t+1}$ determines a discrete move:
\begin{equation}
z^{t+1}=
\begin{cases}
(z^t+1)\bmod d, & r<\sigma(v^{t+1}),\\
(z^t-1)\bmod d, & \text{otherwise},
\end{cases}
\label{eq:pso_discrete}
\end{equation}
where $r\sim U(0,1)$ and $\sigma(v)=(1+e^{-v})^{-1}$. Modulo arithmetic enforces the grid boundaries, and duplicate coordinates are replaced to preserve the budget.

\paragraph{Cuckoo Search (CS)} represents each sampling set as a binary mask and perturbs it through L\'evy flights~\cite{yang2009cuckoo,ouaarab2014discrete,bibiks2015discrete}. The heavy-tailed steps permit long-range exploration of sparse high-frequency regions. We use $\lambda=1.5$ and a step size with a cosine decay from 0.5 to 0.1 over the first 100 iterations, remaining at 0.1 thereafter. Candidate masks remain subject to the sampling budget and coordinate uniqueness. Nest selection and duplicate-point removal refine the resulting sampling sets.

\begin{table}[t]
\centering
\small
\setlength{\tabcolsep}{3pt}
\begin{tabular*}{\columnwidth}{@{\extracolsep{\fill}}cccccc@{}}
\toprule
$f_c$ & BLEU & NIST & METEOR & ROUGE-L & CIDEr\\
\midrule
0   & 68.7 & 8.35 & 41.3 & 67.5 & 1.65\\
100 & 67.3 & 8.23 & 41.2 & 67.8 & \textbf{1.66}\\
200 & 67.6 & \textbf{8.41} & 40.9 & 67.4 & 1.64\\
300 & \textbf{69.2} & 8.36 & \textbf{41.4} & 67.7 & 1.63\\
400 & 68.7 & 8.26 & 40.8 & \textbf{67.9} & 1.64\\
500 & 62.5 & 8.30 & 40.3 & 66.7 & 1.61\\
\bottomrule
\end{tabular*}
\caption{Gaussian band-pass sampling with different frequency biases. Best values are bold.}
\label{tab4}
\end{table}

\section{Experimental Studies}\label{sec4}
We evaluate GPT-2-Medium~\cite{radford2019language} on E2E~\cite{novikova2017e2e}, using 42,000 training examples and 4,600 examples each for validation and testing. Baselines include Full FT, LoRA~\cite{hu2021lora}, LoCA~\cite{du2025loca}, and FourierFT with uniform or Gaussian band-pass sampling~\cite{gao2024parameter}. We report BLEU, NIST, METEOR, ROUGE-L, and CIDEr~\cite{papineni2002bleu,doddington2002nist,banerjee2005meteor,lin2004rouge,vedantam2015cider}.

All PEFT methods adapt the $Q$ and $V$ projections, following common practice for LoRA~\cite{hu2021lora}, while Full FT updates the complete model. All methods use AdamW with batch size 4 for three epochs. Learning rates are $2\times10^{-4}$ for FourierFT and LoCA, $3\times10^{-4}$ for LoRA, and $5\times10^{-5}$ for Full FT. LoRA uses rank 8; LoCA uses 1,000 spectral coefficients per adapted matrix; and FourierFT uses $\alpha=16$. FourierFT and LoCA therefore train 48,000 spectral coefficients across 24 layers and two matrices per layer. HeuFouFT uses 624 selected coordinates and 624 independently trainable real coefficients per matrix, giving $24\times2\times624=29{,}952$ parameters (0.030M), 37.6\% fewer than those configurations. The selected $E$ is fixed across three seeds, so the standard deviations reflect fine-tuning stochasticity rather than search variance.

\begin{table}[t]
\centering
\small
\setlength{\tabcolsep}{1.2pt}
\newcommand{\meanstd}[2]{$\vcenter{\hbox{\shortstack{$#1$\\[1pt]$\pm #2$}}}$}
\newcommand{\methodcell}[1]{$\vcenter{\hbox{\shortstack[l]{#1}}}$}
\begin{tabular*}{\columnwidth}{@{\extracolsep{\fill}}lcccccc@{}}
\toprule
Method & BLEU & NIST & METEOR & ROUGE-L & CIDEr & \shortstack{Params.\\(M)}\\
\midrule
Full FT & 68.9 & 8.36 & 40.1 & 67.2 & 1.59 & 354.92\\\addlinespace[3pt]
LoRA & \meanstd{68.4}{.2} & \meanstd{8.34}{.05} & \meanstd{41.0}{.1} & \meanstd{67.1}{.3} & \meanstd{1.57}{.02} & 0.786\\\addlinespace[3pt]
LoCA & \meanstd{\mathbf{70.3}}{.1} & \meanstd{\mathbf{8.51}}{.03} & \meanstd{\mathbf{41.6}}{.08} & \meanstd{\mathbf{68.5}}{.2} & \meanstd{\mathbf{1.64}}{.01} & 0.048\\\addlinespace[3pt]
\methodcell{Random\\Uniform} & \meanstd{66.7}{.2} & \meanstd{8.24}{.04} & \meanstd{40.2}{.09} & \meanstd{66.5}{.2} & \meanstd{1.54}{.03} & 0.048\\\addlinespace[3pt]
\methodcell{Gaussian\\Band-Pass} & \meanstd{69.2}{.1} & \meanstd{8.36}{.05} & \meanstd{41.4}{.07} & \meanstd{67.7}{.2} & \meanstd{1.63}{.02} & 0.048\\
\midrule
Vanilla GA & \meanstd{69.6}{.1} & \meanstd{8.72}{.03} & \meanstd{41.4}{.08} & \meanstd{67.3}{.2} & \meanstd{1.63}{.01} & 0.03\\\addlinespace[3pt]
GA-SA & \meanstd{70.6}{.1} & \meanstd{8.73}{.03} & \meanstd{41.5}{.07} & \meanstd{68.6}{.2} & \meanstd{1.68}{.01} & 0.03\\\addlinespace[3pt]
PSO & \meanstd{\mathbf{70.9}}{.2} & \meanstd{\mathbf{8.78}}{.02} & \meanstd{\mathbf{41.8}}{.05} & \meanstd{68.5}{.3} & \meanstd{\mathbf{1.71}}{.02} & 0.03\\\addlinespace[3pt]
CS & \meanstd{70.4}{.1} & \meanstd{8.75}{.01} & \meanstd{41.6}{.04} & \meanstd{\mathbf{68.9}}{.2} & \meanstd{1.69}{.01} & 0.03\\\addlinespace[3pt]
\bottomrule
\end{tabular*}
\caption{GPT-2-Medium performance on E2E. Entries with error terms show the mean and standard deviation over seeds 41, 42, and 43. Best baseline and HeuFouFT results are bolded separately.}
\label{tab5}
\end{table}

\subsection{Frequency Bias and Block Sampling}\label{sec4-2}
We test six Gaussian frequency biases, from $f_c=0$ to 500 in steps of 100, with $W=20$ and all other settings fixed. As Table~\ref{tab4} shows, the best setting differs across metrics, so no single $f_c$ places the annulus optimally for all of them. Changing $f_c$ shifts the sampling distribution globally but does not optimize individual in-band coordinates using task feedback. This limitation motivates block-level feedback and coordinate-wise search.

\subsection{Overall Performance}\label{sec4-3}
Table~\ref{tab5} compares the selected coordinate sets with the baselines. Vanilla GA removes simulated annealing from GA-SA, providing an ablation of the acceptance mechanism under the same parameter budget. All three HeuFouFT variants use the same intensity map, so their differences reflect the search strategies under a common prior.

PSO achieves the highest BLEU, NIST, METEOR, and CIDEr, while CS yields the highest ROUGE-L. \textbf{All three HeuFouFT variants outperform the fixed sampling baselines and LoRA.} They do so with 37.6\% fewer coefficients than the FourierFT configurations, showing that task feedback can allocate a smaller spectral budget more effectively than a random-uniform draw or a radial band. Adding simulated annealing to Vanilla GA improves all five metrics. The consistent gain supports its role in escaping local optima that greedy evolution can settle into.

The three optimizers use the common intensity prior in different ways. PSO repeatedly moves candidates toward personal and swarm bests, so its coordinates cluster around a compact middle-frequency region favored by the prior. CS makes occasional long jumps through L\'evy flights and replaces weaker nests, producing broader exploration and more high-frequency coordinates. GA-SA combines crossover and mutation with annealed acceptance of temporary fitness decreases; it therefore preserves a more balanced middle-to-high-frequency distribution and is less likely to collapse around one region. Their similar downstream performance indicates that the task-guided framework supplies the main gain, while the optimizer controls how the region rich in frequency-domain information is explored.

\subsection{Sampling Behavior and Efficiency}\label{sec4-4}
\begin{figure}[t]
\centering
\includegraphics[width=\columnwidth]{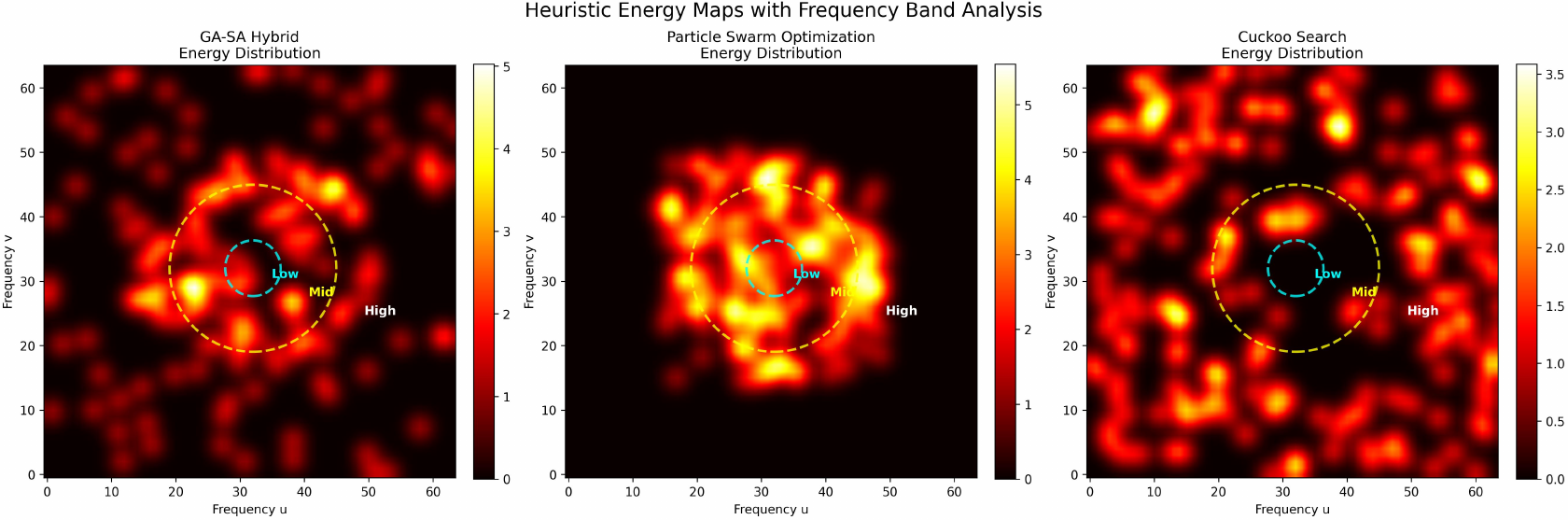}
\caption{Distributions of the selected frequency coordinates, shared across layers, for GA-SA (left), PSO (middle), and CS (right). Dashed circles mark frequency bands.}
\label{figmap}
\end{figure}

\begin{figure}[t]
\centering
\includegraphics[width=\columnwidth]{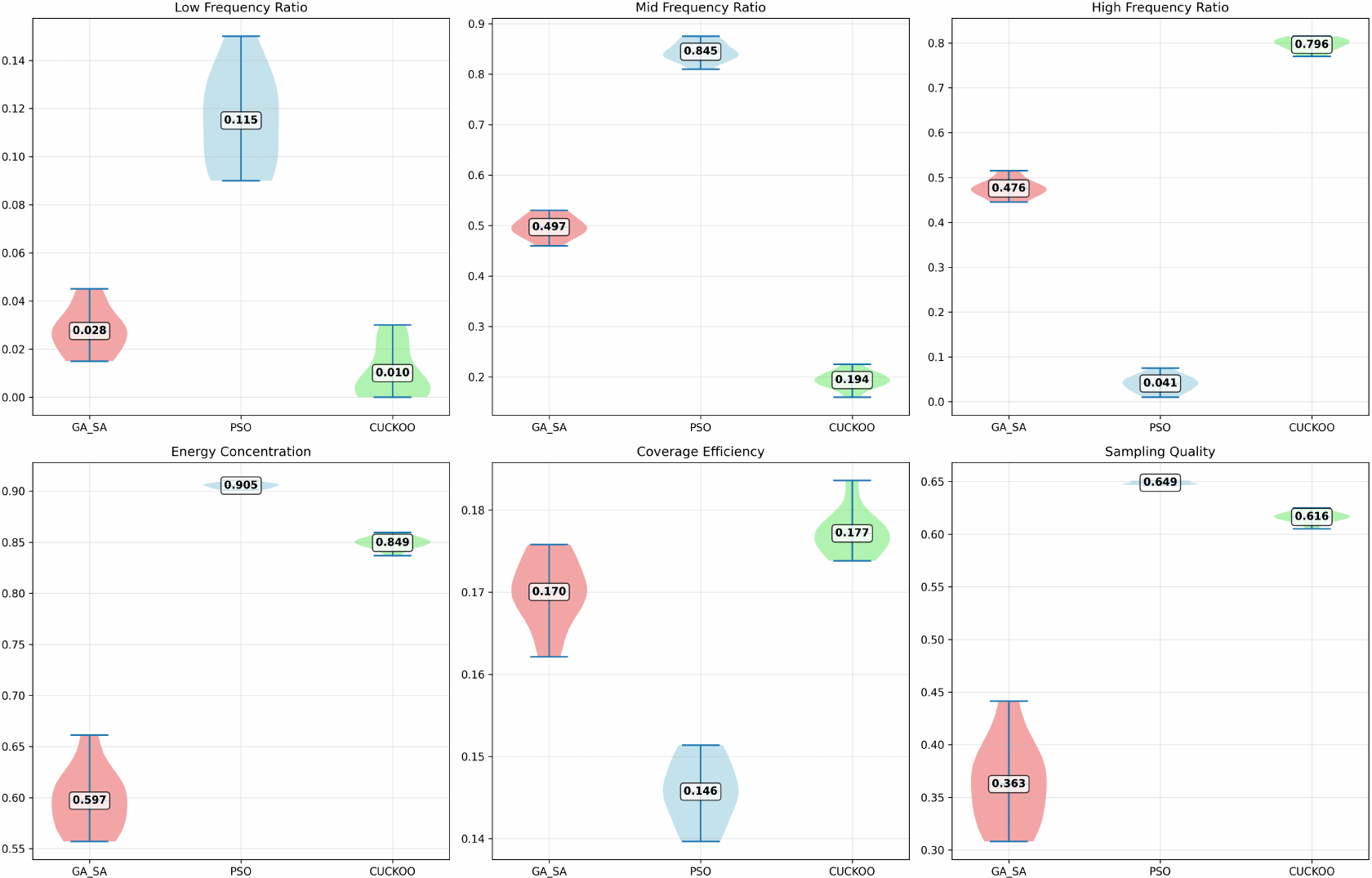}
\caption{Frequency-domain metrics of GA-SA, PSO, and CS. Panels show frequency ratios, energy concentration, coverage efficiency, and sampling quality.}
\label{freq_ana}
\end{figure}

We use four quantities to read Figures~\ref{figmap} and~\ref{freq_ana}. \emph{Energy density} is the normalized squared spectral magnitude at each coordinate. \emph{Energy concentration} is the fraction of total update energy captured by the selected coordinates. \emph{Coverage efficiency} measures how well the selection covers regions that have both high heuristic intensity and high energy density. \emph{Sampling quality} summarizes the balance between concentrating on strong components and covering useful regions; a high value requires both rather than either alone.

The figures confirm the distinct sampling behaviors of the three optimizers. PSO places 84.5\% of its coordinates in the middle band and obtains the highest energy concentration and sampling quality, showing focused exploitation of strong components. CS places 79.6\% in the high band and obtains the highest coverage efficiency, consistent with broader exploration of useful regions. GA-SA splits nearly all coordinates between the middle and high bands and scores lower on the aggregate quality measure, reflecting its more diffuse search. These profiles align with the metric results: PSO leads on four metrics, whereas CS leads on ROUGE-L. The distinction reflects how each optimizer allocates the spectral budget, which enables more flexible frequency-domain coordinate selection.

The Random Forest limits proxy fine-tuning to the top 30\% of each candidate population~\cite{sun2019surrogate}. GA-SA, PSO, and CS finish after 175, 240, and 151 iterations, taking 2.92, 3.00, and 1.26 h, respectively. CS has the shortest search time, while PSO gives the best results on four metrics. Map construction takes 1.25 h; fitting the surrogate, running the search, and final fine-tuning bring the full pipeline to about 12 h on one NVIDIA A100 80GB GPU in FP16. Coordinate selection is paid once for a task and can be amortized when the selected set is reused.

After $E$ is fixed, HeuFouFT uses 15--18\% of Full FT FLOPs and about 9\% of its peak memory. The corresponding figures are 40\% and 28\% for LoRA, and 25\% and 12\% for FourierFT and LoCA. Thus, post-selection HeuFouFT uses 28--40\% fewer FLOPs and 25\% less memory than FourierFT and LoCA; the separate search cost is reported above.

\section{Conclusion}\label{sec5}
We presented HeuFouFT, a task-guided framework that selects Fourier coordinates using block-level probes, Random Forest filtering, and metaheuristic search. On GPT-2-Medium and E2E, all three variants outperform fixed sampling and LoRA across five metrics while using 37.6\% fewer coefficients than the FourierFT and LoCA configurations; PSO further outperforms LoCA on four metrics. Once coordinates are selected, HeuFouFT requires only 15--18\% FLOPs of Full FT. Moreover, the three searches reach similar performance from distinct frequency distributions, so the benefit stems from task-guided selection itself rather than the optimizer. These results demonstrate that \emph{task-guided coordinate search} can allocate limited spectral adaptation capacity effectively.

% Keep all technical content on pages 1--4; the bibliography starts a new page.
\clearpage
\begingroup
\microtypesetup{expansion=true}
\linespread{1}\selectfont
% Preserve the section heading size; use 9 pt only for the reference entries.
\let\icasspOriginalBibliography\thebibliography
\renewcommand{\thebibliography}[1]{%
  \icasspOriginalBibliography{#1}%
  \fontsize{9}{10.4}\selectfont
  \setlength{\itemsep}{0pt}%
  \interlinepenalty=10000\relax
  \setlength{\parsep}{0pt}%
  \setlength{\parskip}{0pt}}
\bibliographystyle{IEEEbib}
\bibliography{references}
\endgroup

% ---- Appendix: included in the arXiv version via \input ----
\clearpage
\appendix
% Appendix body, included by icassp2027.tex via \input after \appendix.
% Preamble dependencies (algpseudocode, tabularx, placeins, makecell, etc.)
% are loaded by the main file.
\section{Frequency Domain Information Metrics}\label{appendixB}
\subsection{Spectral Entropy} 
Spectral entropy is a core indicator for evaluating the uniformity of information distribution in the frequency domain, which is defined as the Shannon entropy of the spectral density function:
\[H_s=-\int_{-\pi}^\pi f_x(\lambda)\log f_x(\lambda)d\lambda \tag{A1}\]

where \(f_x(\lambda)\) is a normalized spectral density function, satisfying \(\int_{-\pi}^{\pi} f_x(\lambda) d\lambda = 1\).

For discrete frequency spectra, the spectral entropy can be approximately calculated as:
\[H_s=-\sum_{u=0}^{M-1}\sum_{v=0}^{N-1}p(u,v)\log p(u,v)\tag{A2}\]

And the normalized spectral energy distribution is as follows:
\[p(u,v)=\frac{|F(u,v)|^2}{\sum\limits_{u^{\prime}=0}^{M-1}\sum\limits_{v^{\prime}=0}^{N-1}|F(u^{\prime},v^{\prime})|^2}\tag{A3}\]

The value range of spectral entropy is \([0, \log(MN)]\). A larger value indicates a more uniform energy distribution in the frequency domain and a lower information density.

\subsection{Spectral Information Density} 
Spectral Information Density (SID) is another important indicator, which is used to measure the information content per unit frequency range. It can be defined by the ratio of spectral entropy to the maximum possible entropy:
\[SID=\frac{H_s}{H_{max}}=\frac{-\sum p(u,v)\log p(u,v)}{\log(MN)}\tag{A5}\]

where \(H_{\text{max}} = \log(MN)\) is the maximum possible entropy value, corresponding to the case of completely uniform distribution.

The value range of spectral information density is \([0, 1]\). A larger value indicates that the information distribution in the frequency domain is closer to a uniform state, and the information density is lower.

\subsection{Low-Frequency Energy Ratio} 
Low-Frequency Energy Ratio (LFER) is another key indicator for evaluating the energy distribution in the frequency domain, which quantifies the proportion of energy in the low-frequency region to the total energy:
\[LFER=\frac{\sum\limits_{(u,v)\in\Omega_L}|F(u,v)|^2}{\sum\limits_{u=0}^{M-1}\sum\limits_{v=0}^{N-1}|F(u,v)|^2}\tag{A6}\]

where \(\Omega_L\) denotes the low-frequency region, which is usually defined as a circular or square region with the spectrum center as the center and radius R.The value range of LFER is \([0, 1]\), and a larger value indicates a higher proportion of low-frequency components in the total energy.
\subsection{Coverage Efficiency}
Coverage Efficiency  measures the weighted coverage of sampling points over regions that are both heuristically critical and energy-dense. It normalizes the weighted count of sampled points in high-value regions by the weighted area of these regions, where ``weight" is defined as the product of Heuristic Intensity (HI) and Energy Density (ED) at each frequency-domain position.
\[
     CE = \frac{\sum_{(u,v)\in S}\mathrm{HI}(u,v)\cdot \mathrm{ED}(u,v)}{\sum_{(u,v)\in \Omega_{\text{eff}}}\mathrm{HI}(u,v)\cdot \mathrm{ED}(u,v)} \cdot \frac{A_{\text{total}}}{A_{\text{eff,weighted}}}
     \tag{A8}
\]

where S: Set of sampled frequency-domain points. \(\text{HI}(u,v)\): Heuristic Intensity at position \((u,v)\), defined as a weighted score of NLG metrics (e.g., \(\text{HI} = 0.3\text{BLEU} + 0.2\text{METEOR} + 0.2\text{ROUGE-L} + 0.15\text{NIST} + 0.15\text{CIDEr}\)) from lightweight fine-tuning of sub-regions. 

\(\text{ED}(u,v)\): Energy Density at position \((u,v)\), defined as \(\text{ED}(u,v) = \frac{|F(u,v)|^2}{\max_{(u',v')}|F(u',v')|^2}\) (normalized frequency-domain coefficient amplitude).\(\Omega_{\text{eff}}\): Set of high-value frequency-domain positions where \(\text{HI}(u,v) \cdot \text{ED}(u,v) \geq \tau\) ( \(\tau\) is a threshold, e.g., the 75th percentile of \(\text{HI} \cdot \text{ED}\) across the entire matrix).\(A_{\text{total}} = M \times N\): Total area of the \(M \times N\) frequency-domain matrix.\(A_{\text{eff, weighted}} = \sum_{(u,v) \in \Omega_{\text{eff}}} \text{HI}(u,v) \cdot \text{ED}(u,v)\): Weighted area of \(\Omega_{\text{eff}}\), capturing both heuristic and energy importance.
\subsection{Energy Concentration}
\label{app:ec_definition}

Energy Concentration (EC) is a core indicator for evaluating the effectiveness of frequency-domain sampling strategies, which quantifies the ability to retain key energy components of the weight update signal ($\Delta W$) during the sampling process. In frequency-domain fine-tuning, the signal energy is typically concentrated in a small number of low-frequency or characteristic frequency components. By calculating the proportion of energy covered by the frequency points selected through sampling, this indicator directly reflects the accuracy of the sampling strategy: a higher EC value indicates that the sampling points are more capable of focusing on core energy components, resulting in less redundant computation and higher fine-tuning efficiency.

Based on the frequency-domain representation of the Discrete Fourier Transform (DFT) and Parseval’s theorem, the mathematical definition of Energy Concentration is given as:
\[
\text{EC} = \frac{\sum\limits_{m=1}^{M} |X_{\text{top-}m}|^2}{\sum\limits_{k=0}^{N-1} |X(k)|^2} \tag{A7}
\label{eq:ec_def}
\]

The detailed explanations of each parameter are as follows:
\begin{enumerate}
    \item $N$: The total number of points in the discrete Fourier transform, corresponding to the length of the frequency-domain sequence of the weight update signal $\Delta W$ after DFT conversion;
    \item $X(k)$: The $k$-th component of the frequency-domain sequence ($k=0,1,\dots,N-1$), where the magnitude $|X(k)|$ reflects the energy intensity of the corresponding frequency component;
    \item $M$: The number of key frequency points selected by the frequency-domain sampling strategy (i.e., the sampling scale of HeuFouFT in this paper, e.g., $M=30000$);
    \item $X_{\text{top-}m}$: The top $M$ frequency-domain components with the highest energy, sorted in descending order of magnitude $|X(k)|$;
    \item The numerator represents the total energy covered by the sampling points, and the denominator represents the full energy of the frequency-domain signal. The value range of EC is $(0,1]$.
\end{enumerate}

\section{NLG Evaluation Metrics} \label{appendixA}
\subsection{BLEU}
BLEU (Bilingual Evaluation Understudy) quantifies the quality of machine-generated text by comparing it to human-annotated reference texts, with scores ranging from 0 to 1 (typically 0.2–0.5 for acceptable performance). Its core principle is to measure the proportion of matching n-grams between candidate and reference texts, with higher overlap indicating better quality. While computationally efficient and widely adopted, BLEU is semantics-agnostic and tends to overvalue short texts. The formula is:
\[BLEU=BP\cdot\exp(\sum_{n=1}^Nw_n\log p_n) \tag{B1}\]

where \(\text{BP} = \exp(1 - L_{\text{ref}}/L_{\text{sys}})\) (length penalty, \(L_{\text{ref}} \geq L_{\text{sys}}\)), \(p_n\) is the clipped n-gram precision, and \(w_n\) are weights (default: uniform for 1–4-grams).

\subsection{NIST}
NIST (National Institute of Standards and Technology) improves BLEU by incorporating information-weighted n-grams (via mutual information entropy). Unlike BLEU’s geometric mean, NIST uses a weighted arithmetic mean, emphasizing rare but informative n-grams. Scores range from 0 to 10 (2–8 considered ideal). Advantages include word order and semantic sensitivity, though it underweights low-frequency terms. The formula is:
\[\mathrm{NIST}=\frac{1}{N}\sum\limits_{n=1}^N\frac{\sum\limits_{t\in T}c_t\log\frac{c_t}{\left|Ref\right|}}{\sum\limits_{t\in T}c_t} \tag{B2}\]

where T is the set of n-grams in references, \(c_t\) is the clipped count in the candidate, and \(\beta_n\) balances n-gram orders.

\subsection{METEOR}

METEOR (Metric for Evaluation of Translation with Explicit Ordering) aligns candidate and reference texts via stem/synonym matching and penalizes word-order mismatches. It computes a harmonic mean of precision (P) and recall (R), with parameters $\alpha,\beta,\gamma$ (default: 0.9, 3, 0.5). Scores range from 0 to 1 (0.3–0.6 for good quality). Strong semantic sensitivity and word-order awareness come at the cost of higher computational complexity. The formula is:
\[\mathrm{METEOR}=\frac{1}{1+\alpha\cdot\beta^\gamma}\cdot\frac{1+\gamma\cdot F1}{1+\gamma}\tag{B3}\]

\subsection{ROUGE-L}
ROUGE-L (Recall-Oriented Understudy for Gisting Evaluation–Longest Common Subsequence) measures similarity via the longest common subsequence (LCS) between candidate and reference summaries, capturing paraphrasing and semantic coherence. Scores range from 0 to 1 (0.2–0.4 typical), with $\beta$ balancing recall (R) and precision (P). Advantages include long-range semantic capture, though word-order details are ignored. The formula is:
\[\mathrm{ROUGE-L}=\frac{(1+\beta^2)\cdot R_{\mathrm{LCS}}\cdot P_{\mathrm{LCS}}}{\beta^2\cdot R_{\mathrm{LCS}}+P_{\mathrm{LCS}}}\tag{B4}\]
\[R_{\mathrm{LCS}}=\frac{\mathrm{LCS}(X,Y)}{|X|} \tag{B5} \]

\[P_{\mathrm{LCS}}=\frac{\mathrm{LCS}(X,Y)}{|Y|}\tag{B6}\]

where X is the reference, Y the candidate, and $\beta \geq$ 1 (default: $\infty$ for recall emphasis).

\subsection{CIDEr}
CIDEr (Consensus-based Image Description Evaluation) measures coherence via TF-IDF-weighted n-gram similarity between candidate and reference descriptions, with geometric averaging across references. Scores range from 0 to 10 ($\geq$1.0 favorable), rewarding semantic diversity while ignoring syntax errors. The formula is:
\[\mathrm{CIDEr}=\exp(\frac{1}{M}\sum_{m=1}^M\log sim_m)\tag{B7}\]

where M is the number of references, \(\text{sim}_m\) is the TF-IDF cosine similarity with reference \(r_m\).

\section{Random Forest Surrogate Model for Meta-Heuristic Frequency-Domain Sampling}
\label{app:rf_surrogate}

\subsection{Motivation and Design Rationale}
The Random Forest (RF) surrogate model in HeuFouFT is designed to filter low-potential frequency candidates prior to meta-heuristic optimization (GA-SA, PSO, CS), reducing redundant computations while preserving high-performance sampling points. Its core design aligns with HeuFouFT's heuristic sampling philosophy: \textbf{prioritize frequency components that contribute most to fine-tuning performance} by leveraging multi-dimensional spectral features. The model is trained to map spectral characteristics of frequency candidates to their actual fine-tuning performance (e.g., BLEU score), enabling rapid screening of high-value candidates.

\subsection{Feature Selection for Spectral Characterization}
Different from the frequency evaluation measures of Appendix A, we select 7 spectral features to capture the energy distribution, information density, and frequency structure of candidate frequency points—covering mandatory indicators easy to calculate and domain-specific supplementary features in table~\ref{tab:rf_features}. All features are computed based on the DFT-transformed weight update signal $\Delta W$ (denoted as $X(k)$ for the $k$-th frequency component, $k=0,1,\dots,N-1$).

% 光谱特征表：[!t]强制定位顶部，与“特征选择”内容紧密衔接

\subsection{Hyperparameter Configuration}
The RF hyperparameters are tuned to balance prediction accuracy and computational efficiency, consistent with HeuFouFT's experimental setup (GPT-2 Medium, E2E NLG dataset) and meta-heuristic optimization constraints. Key hyperparameters and their optimal values are summarized in Table \ref{tab:rf_hyperparams}:

\subsection{Training Algorithm Flow}
The RF surrogate model training follows a data-driven pipeline integrated with HeuFouFT's heuristic sampling framework, as detailed in Algorithm \ref{alg:rf_training}:

\subsection{Computational Cost Evaluation}
We evaluate the computational cost of the RF surrogate model on a single NVIDIA A100 80GB GPU (FP16 precision), consistent with HeuFouFT's experimental hardware. The cost comparison (vs. direct meta-heuristic optimization) is shown in Table \ref{tab:rf_cost}, placed at the top of the page for easy reference:

Key observations:
1. The RF surrogate model reduces overall computational cost by ~68\% by filtering 70\% of low-potential candidates.
2. Data collection dominates the RF-related cost (81\%), which is a one-time overhead for multiple meta-heuristic iterations.
3. The inference stage is computationally lightweight (less than 0.5 GPU-hours), ensuring minimal latency in the sampling pipeline.
4. The cost aligns with typical LLM fine-tuning expenses on cloud GPUs, making the approach scalable for larger models (e.g., Llama 3-7B).

\section{Computational Cost Analysis of Meta-Heuristic Sampling Algorithms}
\label{app:computational_cost}
This section details the exact compute budget, wall-clock time, and measurement methodology for key components of HeuFouFT. All experiments are conducted on a single NVIDIA A100 80GB GPU with FP16 precision, using PyTorch 2.1 and CUDA 12.2 for acceleration.

For constructing the heuristic intensity map, the $1024 \times 1024$ frequency-domain matrix is partitioned into 100 non-overlapping $10 \times 10$ blocks, with each block undergoing lightweight fine-tuning (10\% of E2E training data: 4,200 samples; batch size=4; max\_seq\_len=128; 1 epoch) to compute heuristic intensity via weighted NLG metrics (BLEU, METEOR, ROUGE-L, NIST, CIDEr). Each block takes 45 seconds for forward/backward passes and metric computation, leading to a total wall-clock time of $100 \times 45\ \text{s} = 75\ \text{minutes}$ (~1.25 hours) and total FLOPs of $8.2 \times 10^{17}$ (with $8.2 \times 10^{15}$ FLOPs per block). For labeling 5000 schemes for the surrogate model, each randomly generated sampling scheme is evaluated via the same lightweight fine-tuning setup (no parameter updates), taking 30 seconds per scheme to obtain heuristic intensity labels; this results in a total wall-clock time of $5000 \times 30\ \text{s} = 25,000\ \text{s}$ (~6.94 hours) and total FLOPs of $2.05 \times 10^{19}$ (with $4.1 \times 10^{15}$ FLOPs per scheme).

Regarding the meta-heuristic searches, the GA-SA hybrid algorithm uses a population size of 100, 175 total iterations, a mutation probability of 0.05, initial temperature of 1000, and decay rate of 0.95, with per-iteration operations including two-point crossover, bit-flip mutation, and Metropolis acceptance criterion; each iteration takes 60 seconds, leading to a total wall-clock time of $175 \times 60\ \text{s} = 175\ \text{minutes}$ (~2.92 hours) and total FLOPs of $6.3 \times 10^{16}$ (with $3.6 \times 10^{14}$ FLOPs per iteration). 

The Particle Swarm Optimization (PSO) algorithm employs 50 particles, 240 total iterations, a linearly decreasing inertia weight (from 0.9 to 0.4), and acceleration constants $c_1 = c_2 = 1.4945$, with per-iteration operations including velocity update, sigmoid-based position adjustment, and boundary constraint enforcement via modulo arithmetic; each iteration takes 45 seconds, resulting in a total wall-clock time of $240 \times 45\ \text{s} = 180\ \text{minutes}$ (3 hours) and total FLOPs of $6.72 \times 10^{16}$ (with $2.8 \times 10^{14}$ FLOPs per iteration). 

The Cuckoo Search (CS) with Lévy Flights uses 20 nests, 151 total iterations, a Lévy flight parameter $\lambda = 1.5$, a cosine-decayed step size $\alpha$ (from 0.5 to 0.1), and a nest elimination probability of 0.25, with per-iteration operations including Lévy flight perturbation, nest selection, and duplicate point removal; each iteration takes 30 seconds, leading to a total wall-clock time of $151 \times 30\ \text{s} = 75.5\ \text{minutes}$ (~1.26 hours) and total FLOPs of $2.87 \times 10^{16}$ (with $1.9 \times 10^{14}$ FLOPs per iteration).

The FLOP and memory reduction metrics reported in the main text are end-to-end, encompassing all steps of HeuFouFT: heuristic intensity map construction, surrogate model labeling and training, meta-heuristic sampling search, and final fine-tuning of the LLM, ensuring a comprehensive comparison with baselines (e.g., LoRA, vanilla FourierFT) that only include final training. For FLOPs calculation, we track floating-point operations via \texttt{torch.cuda.flops\_counter()} across all computational graphs (forward/backward passes for fine-tuning, sampling scheme generation, surrogate model training), with FP16 precision consistently applied to align with GPU hardware capabilities and reduce redundant computations. Peak GPU memory usage is recorded via \texttt{torch.cuda.max\_memory\_allocated()}, including model weights (LLM + surrogate), intermediate activations, sampling scheme storage, and gradient buffers, with cache management (e.g., clearing unused tensors post-iteration) applied to mimic real-world deployment. Wall-clock time is measured via \texttt{time.perf\_counter()}, with data loading time excluded (preprocessed E2E data is cached in GPU memory to eliminate I/O overhead), and each step is repeated 3 times to report the average time accounting for minor runtime fluctuations.

% 跨栏伪代码（无多余数字、无语法错误）
\begin{algorithm*}[!t]

\caption{Random Forest Surrogate Model Training for HeuFouFT}
\label{alg:rf_training}
\begin{algorithmic}[1]  % 改用algorithmic环境（兼容宏包，解决未定义报错），[1]显示行号
\Require Weight update signal $\Delta W \in \mathbb{R}^{D \times D}$, DFT size $N$, candidate sampling size $S$, meta-heuristic sampling scale $M$
\Ensure Trained RF surrogate model $f_{\text{RF}}$, high-potential frequency candidate set $\mathcal{C}_{\text{high}}$

\State \textbf{Step 1: Training Data Collection}
\State Randomly sample $S=5000$ frequency candidate sets $\{\mathcal{C}_1, \mathcal{C}_2, \dots, \mathcal{C}_S\}$ from the frequency domain $\{0,1,\dots,N-1\}$, where each $\mathcal{C}_i$ has size $M=30000$
\For{each $\mathcal{C}_i$}
    \State Perform FourierFT fine-tuning with $\mathcal{C}_i$ to get performance label $y_i$ (BLEU score)
    \State Compute 7 spectral features (Section \ref{app:rf_surrogate} C.2) for $\mathcal{C}_i$ to form feature vector $\mathbf{x}_i$
\EndFor
\State Construct training dataset $\mathcal{D} = \left\{(\mathbf{x}_i, y_i)\right\}_{i=1}^S$

\State \textbf{Step 2: Dataset Preprocessing}
\State Split $\mathcal{D}$ into training set $\mathcal{D}_{\text{tr}}$ (80\%) and validation set $\mathcal{D}_{\text{val}}$ (20\%)
\State Normalize feature vectors in $\mathcal{D}_{\text{tr}}$ and $\mathcal{D}_{\text{val}}$ using min-max scaling

\State \textbf{Step 3: Hyperparameter Optimization}
\State Use Bayesian Optimization to search hyperparameter space (Table \ref{tab:rf_hyperparams}) to maximize $R^2$ on $\mathcal{D}_{\text{val}}$

\State \textbf{Step 4: Model Training and Validation}
\State Train RF model with optimized hyperparameters on $\mathcal{D}_{\text{tr}}$
\State Evaluate model performance on $\mathcal{D}_{\text{val}}$; if $R^2 < 0.8$, augment $\mathcal{D}$ with 1000 additional samples and retrain

\State \textbf{Step 5: High-Potential Candidate Selection}
\State Apply $f_{\text{RF}}$ to predict performance of all candidate sets generated by initial random sampling
\State Select top 30\% candidates with highest predicted performance to form $\mathcal{C}_{\text{high}}$
\State Output $\mathcal{C}_{\text{high}}$ for subsequent meta-heuristic optimization (GA-SA/PSO/CS)
\end{algorithmic}  % 对应修改结束环境，改为algorithmic
\end{algorithm*}

% 超参数配置表：[!t]强制定位顶部，补充原代码缺失的内容
\begin{table*}[!t]
\centering
\small
\caption{Hyperparameter Configuration of Random Forest Surrogate Model}
\label{tab:rf_hyperparams}
\begin{tabularx}{\linewidth}{lcc>{\raggedright\arraybackslash}X}
\toprule
\textbf{Hyperparameter} & \textbf{Value Range} & \textbf{Optimal Value} & \textbf{Design Rationale} \\
\midrule
Number of Estimators & 100–500 & 300 & Balances model variance (fewer trees: high variance) and computational cost (more trees: high cost). \\
\addlinespace
Max Depth & 10–50 & 25 & Prevents overfitting (too deep: memorizes noise) while capturing complex feature-performance relationships. \\
\addlinespace
Min Samples Split & 2–10 & 5 & Ensures splits are statistically meaningful (avoids trivial splits on small subsets). \\
\addlinespace
Min Samples Leaf & 1–5 & 2 & Reduces overfitting by requiring minimum samples in leaf nodes (avoids isolated data points). \\
\addlinespace
Max Features & $\sqrt{d}$–$d$ ($d=7$) & $\sqrt{d}$ & Uses square-root of total features to balance diversity (randomness) and predictive power. \\
\addlinespace
Learning Rate (for boosting) & 0.01–0.1 & — & Not used (RF is an ensemble of unboosted trees). \\
\bottomrule
\end{tabularx}
\vspace{0.2em}
\footnotesize{Hyperparameters are optimized via Bayesian Optimization (Section \ref{app:rf_surrogate} C.4) to maximize $R^2$ on the validation set.}
\end{table*}

% 计算成本表：[!t]强制定位顶部，与“成本评估”内容衔接
\begin{table*}[!t]
\centering
\small
\caption{Computational Cost Comparison}
\label{tab:rf_cost}
\begin{tabularx}{\linewidth}{>{\raggedright\arraybackslash}Xccc}
\toprule
\textbf{Computational Stage} & \textbf{Wall-Clock Time} & \textbf{GPU-Hours} & \textbf{Cost (USD, 10 USD/hour)} \\
\midrule
RF Training Data Collection & 12.0 hours & 12.0 & 120.0 \\
\addlinespace
RF Model Training (incl. hyperparameter tuning) & 2.5 hours & 2.3 & 23.0 \\
\addlinespace
High-Potential Candidate Selection (Inference) & 0.3 hours & 0.3 & 3.0 \\
\addlinespace
\textbf{Total RF-Related Cost} & \textbf{14.8 hours} & \textbf{14.6} & \textbf{146.0} \\
\addlinespace
Direct Meta-Heuristic Optimization (No RF) & 48.0 hours & 45.2 & 452.0 \\
\addlinespace
\textbf{Reduction Ratio} & \textbf{69.2\%} & \textbf{67.7\%} & \textbf{67.7\%} \\
\bottomrule
\end{tabularx}
\vspace{0.2em}
\end{table*}

\FloatBarrier
\begin{table*}[!ht]
\centering
\small
\caption{Spectral Features for Random Forest Surrogate Model}
\label{tab:rf_features}
\begin{tabularx}{\linewidth}{ll>{\raggedright\arraybackslash}X>{\raggedright\arraybackslash}X}
\toprule
\textbf{Feature Name} & \textbf{Abbreviation} & \textbf{\makecell{Mathematical\\Definition}} & \textbf{\makecell{Physical\\Interpretation}} \\
\midrule
Spectral Entropy & SE & $SE = -\sum\limits_{k=0}^{N-1} p_k \log_2 p_k$, $p_k = \frac{|X(k)|^2}{\sum\limits_{i=0}^{N-1} |X(i)|^2}$ & Measures the uniformity of energy distribution; lower values indicate concentrated energy. \\
\addlinespace
Spectral Information Density & SID & $SID = \frac{1}{N} \sum\limits_{k=0}^{N-1} |X(k)|^2 \cdot \log_2 (|X(k)|^2 + \epsilon)$ & Quantifies information content per frequency component; higher values mean richer task-relevant information. \\
\addlinespace
Low/Medium/High Frequency Ratio & LMH-FR & $LMH\text{-}FR = \sum |X_{\text{low}}|^2 \colon \sum |X_{\text{med}}|^2 \colon \sum |X_{\text{high}}|^2$ & Ratio of energy in three frequency bands (low: $k=0\sim N/3$, med: $N/3\sim 2N/3$, high: $2N/3\sim N-1$). \\
\addlinespace
Energy Concentration & EC & $EC = \frac{\sum\limits_{m=1}^{M} |X_{\text{top-}m}|^2}{\sum\limits_{k=0}^{N-1} |X(k)|^2}$ & Proportion of energy covered by top-$M$ high-energy components (Section \ref{app:ec_definition}). \\
\addlinespace
Frequency Band Coverage & FBC & $FBC = \frac{f_{\text{max}}}{f_{\text{min}}}$, $f_{\text{min/max}}$ = min/max frequency of candidate set & Indicates the span of frequency bands; reflects coverage of spectral characteristics. \\
\addlinespace
Spectral Crest Factor & SCF & $SCF = \frac{\max(|X(k)|^2)}{\frac{1}{N}\sum\limits_{k=0}^{N-1} |X(k)|^2}$ & Measures the concentration of energy in a single frequency component; higher values mean stronger peak energy. \\
\addlinespace
Peak-to-Average Power Ratio & PAPR & $PAPR = 10\log_{10}\left(\frac{\max(|X(k)|^2)}{\mathbb{E}[|X(k)|^2]}\right)$ & Quantifies the discrepancy between peak and average power; correlates with signal stability. \\
\bottomrule
\end{tabularx}
\vspace{0.2em}
\footnotesize{$\epsilon = 10^{-8}$ is added to avoid numerical instability in logarithmic operations. All features are normalized to $[0,1]$ using min-max scaling.}
\end{table*}

\end{document}